\documentclass[10pt,twocolumn,letterpaper]{article}

\usepackage[pagenumbers]{cvpr} 

\usepackage{multirow}
\usepackage{booktabs}
\usepackage{listings}
\usepackage{amsmath}
\usepackage{amsfonts}
\usepackage{algorithm}
\usepackage{algorithmic}

\definecolor{cvprblue}{rgb}{0.21,0.49,0.74}
\usepackage[pagebackref,breaklinks,colorlinks,allcolors=cvprblue]{hyperref}

\def\paperID{3113} 
\def\confName{CVPR}
\def\confYear{2025}

\title{UniVAD: A Training-free Unified Model for Few-shot Visual Anomaly Detection}

\author{
Zhaopeng Gu$^{1,2}$~~~~
Bingke Zhu$^{1,3}$~~~~
Guibo Zhu$^{1,2}$~~~~\\
Yingying Chen$^{1,3}$~~~~
Ming Tang$^{1,2}$~~~~
Jinqiao Wang$^{1,2,3}$\\
  $^{1}$~Foundation Model Research Center, Institute of Automation, \\ Chinese Academy of Sciences, Beijing, China \\ 
  $^{2}$~University of Chinese Academy of Sciences, Beijing, China\\
  $^{3}$~Objecteye Inc., Beijing, China\\
  {\tt\small  guzhaopeng2023@ia.ac.cn} \\
  {\tt\small \{bingke.zhu,gbzhu,yingying.chen,tangm,jqwang\}@nlpr.ia.ac.cn} \\[0.5pt]
  \href{https://uni-vad.github.io}{https://uni-vad.github.io} \\
}

\begin{document}
\maketitle
\begin{abstract}
Visual Anomaly Detection (VAD) aims to identify abnormal samples in images that deviate from normal patterns, covering multiple domains, including industrial, logical, and medical fields. Due to the domain gaps between these fields, existing VAD methods are typically tailored to each domain, with specialized detection techniques and model architectures that are difficult to generalize across different domains. Moreover, even within the same domain, current VAD approaches often follow a ``one-category-one-model" paradigm, requiring large amounts of normal samples to train class-specific models, resulting in poor generalizability and hindering unified evaluation across domains. To address this issue, we propose a generalized few-shot VAD method, UniVAD, capable of detecting anomalies across various domains, such as industrial, logical, and medical anomalies, with a training-free unified model. UniVAD only needs few normal samples as references during testing to detect anomalies in previously unseen objects, without training on the specific domain. Specifically, UniVAD employs a \textbf{C}ontextual \textbf{C}omponent \textbf{C}lustering~($\text{C}^3$) module based on clustering and vision foundation models to segment components within the image accurately, and leverages \textbf{C}omponent-\textbf{A}ware \textbf{P}atch \textbf{M}atching~(CAPM) and \textbf{G}raph-\textbf{E}nhanced \textbf{C}omponent \textbf{M}odeling~(GECM) modules to detect anomalies at different semantic levels, which are aggregated to produce the final detection result. We conduct experiments on nine datasets spanning industrial, logical, and medical fields, and the results demonstrate that UniVAD achieves state-of-the-art performance in few-shot anomaly detection tasks across multiple domains, outperforming domain-specific anomaly detection models. Code is available at https://github.com/FantasticGNU/UniVAD.
\end{abstract}
    
\section{Introduction}

\begin{figure}[t]
  \centering
   \includegraphics[width=0.95\linewidth]{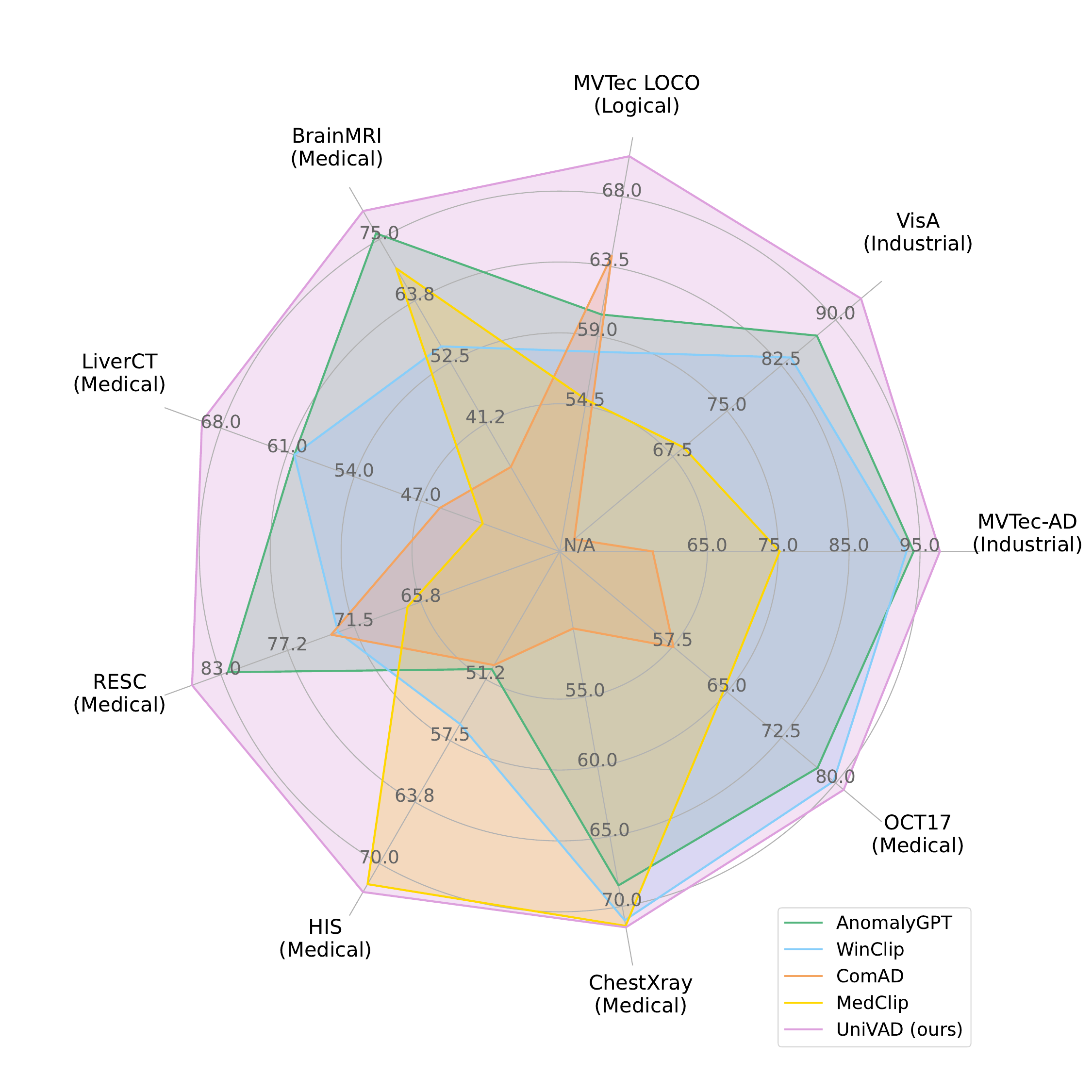}
   \vspace{1mm}
   \caption{1-shot performance of existing VAD methods and UniVAD across different datasets in various domains. UniVAD achieves state-of-the-art results across multiple datasets and domains, outperforming specialized methods in each domain.}
   \label{fig:radar}
\end{figure}

Visual Anomaly Detection (VAD)~\cite{roth2022towards, gu2024anomalygpt, jeong2023winclip, you2022unified} is a critical task that seeks to identify abnormal samples in images that deviate from established normal patterns, leveraging computer vision techniques~\cite{radford2021learning, he2016deep}. Such anomalies are rare occurrences but signify critical conditions, including errors, defects, or lesions, that necessitate timely intervention for further analysis. VAD spans multiple fields and has applications across diverse industries, such as industrial anomaly detection~\cite{roth2022towards, jeong2023winclip, gu2024anomalygpt}, logical anomaly detection~\cite{hsieh2024csad, liu2023component, kim2024few}, and medical anomaly detection~\cite{huang2024adapting, bao2024bmad}.

However, significant variations in data distributions and anomaly types across domains result in current VAD methods~\cite{roth2022towards, liu2023component, bao2024bmad} being highly specialized for specific domains, often employing custom detection algorithms and model architectures. Consequently, methods optimized for one domain tend to perform poorly in others. For instance, one of the state-of-the-art industrial anomaly detection methods, PatchCore~\cite{roth2022towards}, achieves a 1-shot image-level AUC of 84.1\% on the industrial dataset MVTec-AD~\cite{bergmann2019mvtec}. However, its performance drops significantly to 62.0\% when applied to the MVTec LOCO~\cite{bergmann2022beyond} dataset for logical anomaly detection. Furthermore, even within the same domain, most contemporary VAD approaches adopt a ``one-category-one-model" paradigm, where a separate model is trained for each object category. Once trained, each model is limited to the specific object category. This domain- and category-specific approach constrains VAD research's standardization and scalability.

To address these limitations, we propose a training-free generalized anomaly detection method, UniVAD, which leverages a unified model to detect anomalies across multiple domains. UniVAD can handle industrial, logical, medical, and other anomalies without requiring domain-specific data training. Instead, UniVAD requires only a few normal samples of the target category during the testing phase to perform anomaly detection. This approach significantly enhances the generalizability and transferability of anomaly detection models, as illustrated in Figure~\ref{fig:radar}.

Specifically, UniVAD utilizes a contextual component clustering module, incorporating clustering techniques~\cite{liu2023component} and visual foundation models~\cite{huang2023open, kirillov2023segment, liu2023grounding} to segment components within an image. Following this, UniVAD applies a component-aware patch-matching module and a graph-enhanced component modeling module to detect anomalies at various semantic levels. The component-aware patch-matching module identifies anomalies such as structural defects or tissue lesions by matching patch-level features within each component. Meanwhile, the graph-enhanced component modeling module employs graph-based component feature aggregation to model relationships between image components, facilitating the detection of more complex logical anomalies, such as missing, added, or incorrect components, through inter-component feature matching.

Our experiments, conducted across nine datasets covering industrial, logical, and medical domains, such as MVTec-AD~\cite{bergmann2019mvtec}, VisA~\cite{zou2022spot}, MVTec LOCO~\cite{bergmann2022beyond}, Brain MRI~\cite{baid2021rsna}, and Liver CT~\cite{bilic2023liver, landman2015miccai}, demonstrate that UniVAD achieves state-of-the-art performance in few-shot anomaly detection across multiple domains, significantly outperforming domain-specific models.

\begin{figure}[t]
  \centering
   \includegraphics[width=\linewidth]{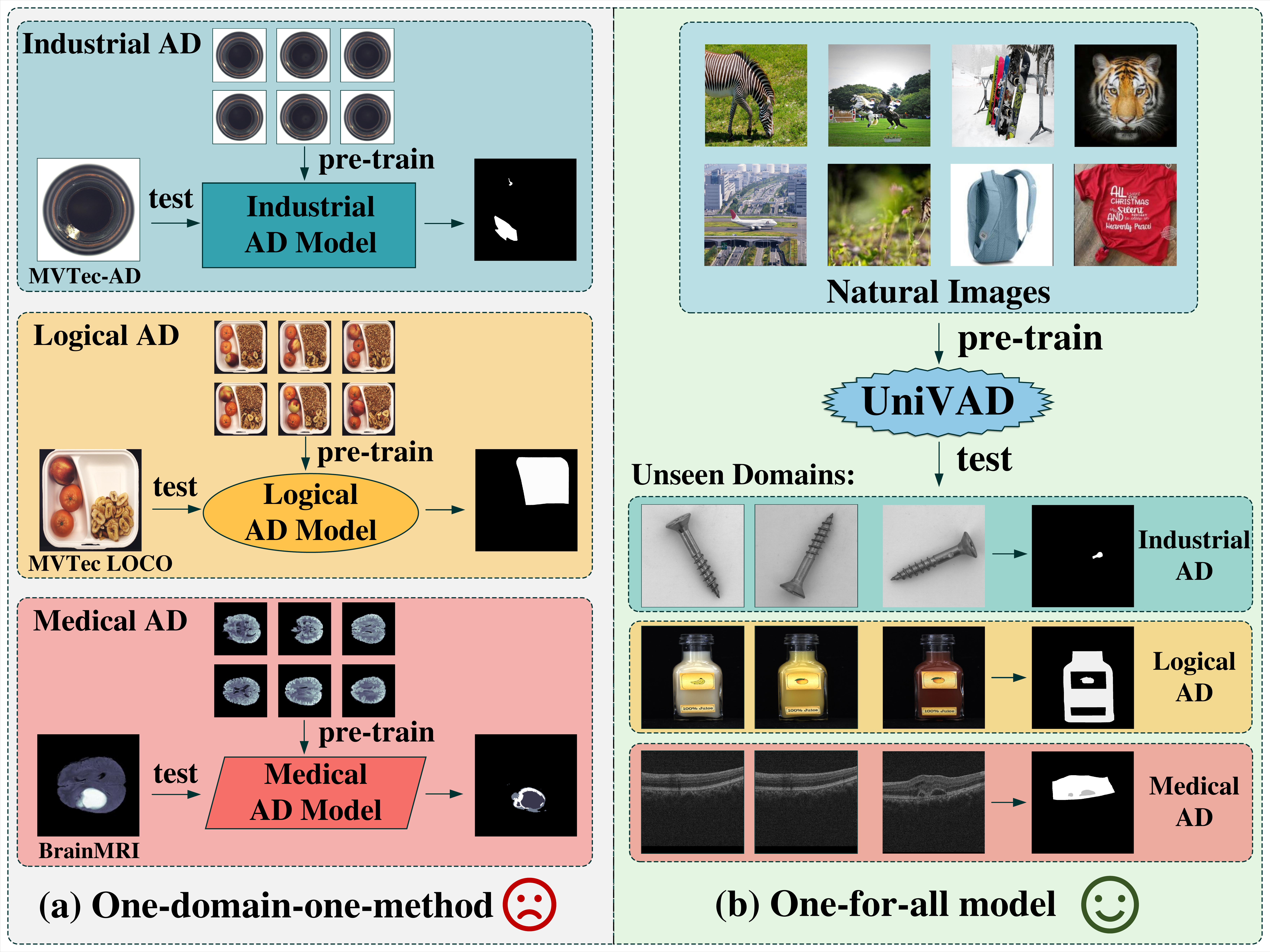}
   \caption{Comparison between UniVAD and existing VAD methods. Existing VAD methods are specifically designed for each domain, whereas UniVAD can perform anomaly detection tasks across multiple domains using a unified model.}
   \label{fig:compare}
\end{figure}

Our contributions are listed as follows:

\begin{itemize}
    \item We introduce the first training-free unified few-shot visual anomaly detection method capable of detecting anomalies across industrial, logical, and medical domains. This unified approach reduces the significant workload to develop separate detection methods and model architectures for each domain, thereby promoting the standardization of anomaly detection research.
    
    \item We design a contextual component clustering module that effectively segments object components under few-shot learning conditions. Additionally, combined with our component-aware patch matching module and graph-enhanced component modeling module, UniVAD reliably detects anomalies across different semantic levels. 
    
    \item Comprehensive experiments on nine datasets spanning industrial, logical, and medical domains demonstrate that UniVAD achieves state-of-the-art performance in few-shot anomaly detection, establishing its effectiveness and generalizability across domains. \\[1mm]
\end{itemize}

\section{Related Work}

\subsection{Visual Anomaly Detection}

In traditional visual anomaly detection, methods are tailored to specific domains to accommodate distinct data characteristics. Industrial anomaly detection focuses on identifying defects, which are generally small and localized, prompting recent methods to emphasize local image features. Practical approaches include patch feature matching, where patch features of test samples are compared to those of normal samples to compute anomaly scores~\cite{roth2022towards, chen2023april}, and reconstruction-based methods that leverage networks trained on normal samples to identify anomalies through reconstruction loss~\cite{hyun2024reconpatch, you2022unified}. Additionally, other approaches employ pre-trained models, such as CLIP~\cite{radford2021learning}, to evaluate patch features against textual descriptions of ``normal" and ``anomalous" states, facilitating a versatile approach to anomaly detection~\cite{jeong2023winclip, gu2024filo}.

Logical anomaly detection~\cite{bergmann2022beyond} assesses whether an image adheres to logical constraints, such as correct components, colors, or quantities, requiring a higher level of semantic understanding. Logical anomalies typically result from incorrect combinations of normal elements. Methods in this area often involve segmenting components~\cite{hsieh2024csad, liu2023component, peng2024sam} to evaluate individual features such as color, area, and quantity, thereby ensuring logical coherence.

Medical anomaly detection aims to locate pathological regions in medical images. Approaches in this domain include  GAN-based~\cite{han2021madgan}, reconstruction-based~\cite{cai2024rethinking}, and self-supervised learning methods~\cite{tian2023self}. However, variability across different body parts and diseases continues to pose substantial challenges for generalizability.

Recent efforts aim to develop unified models for anomaly detection. For example, UniAD~\cite{you2022unified} employs a patch feature reconstruction method optimized for industrial applications; however, its performance decreases in other domains and requires large volumes of normal data samples for model training.

In contrast, UniVAD operates across domains using a training-free unified model and requires only a few normal samples for reference during testing, as shown in Figure~\ref{fig:compare}. This approach eliminates the need for prior training on domain-specific data, offering greater flexibility across various anomaly detection tasks.

\begin{figure*}[t]
  \centering
   \includegraphics[width=\textwidth]{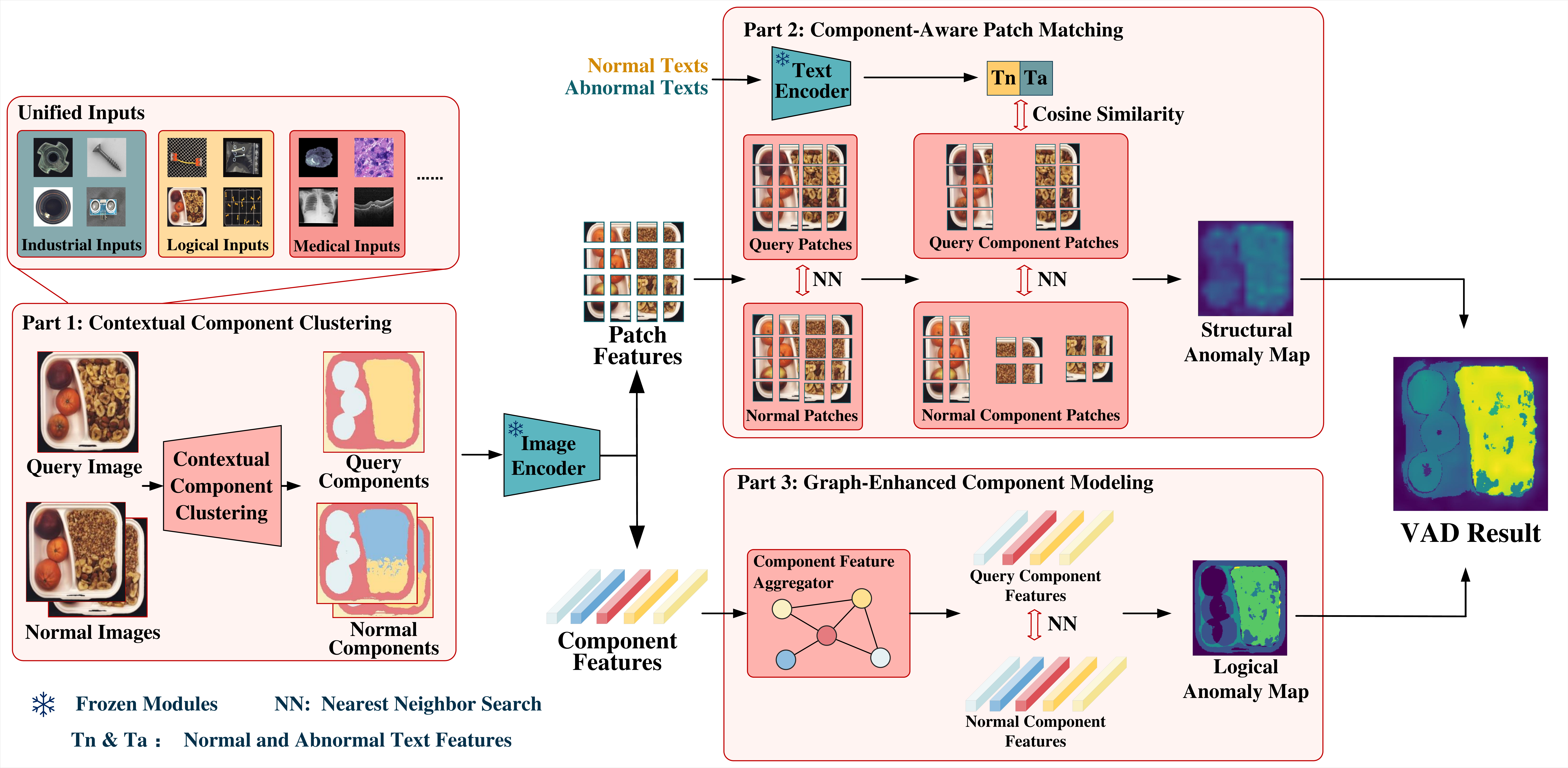}
   \caption{The overall architecture of UniVAD. Given an input image, UniVAD first generates masks for each entity using the Contextual Component Clustering module (Sec 3.2). UniVAD then applies the Component-Aware Patch Matching module (Sec 3.3) and the Graph-Enhanced Component Modeling module (Sec 3.4) to detect structural and logical anomalies. The outputs from both expert modules are combined to produce the final unified anomaly detection result.}
   \label{fig:arch}
\end{figure*}

\subsection{Component Segmentation}

Logical anomaly detection frequently relies on component segmentation to extract sub-parts of an image and evaluate each for anomalies. ComAD~\cite{liu2023component} introduces this approach by using clustering for segmentation; however, clustering requires many samples, limiting its applicability in few-shot scenarios. More recent methods, such as CSAD~\cite{hsieh2024csad} and SAM-LAD~\cite{peng2024sam}, leverage vision models like SAM~\cite{kirillov2023segment} but face challenges with segmentation granularity, often producing outputs that are either overly fine or too coarse. PSAD~\cite{kim2024few} addresses this issue by employing a limited set of annotated samples, which increases training costs and requires manual labeling.

UniVAD combines clustering with vision foundation models, utilizing vision foundation models to produce initial component masks and refining segmentation granularity through clustering. This approach allows for precise segmentation in few-shot settings, enhancing the model's ability to perform effectively with minimal data.

\section{Method}

\subsection{Overall Architecture}
Given a query image $I_q \in \mathbb{R}^{H \times W \times 3}$ and $K$ reference normal images $I_n \in \mathbb{R}^{K \times H \times W \times 3}$, UniVAD first employs a contextual component clustering module to segment components within both the query and reference images, producing corresponding component masks. UniVAD then utilizes an image encoder pre-trained on large-scale datasets to extract features from the query and normal images, resulting in a feature map $F_q \in \mathbb{R}^{H_1 \times W_1 \times C}$ for the query image and $F_n \in \mathbb{R}^{K \times H_1 \times W_1 \times C}$ for the normal images. By applying group average pooling based on the component masks, UniVAD obtains component-level features for both the query and normal images, denoted as $F_{qc} \in \mathbb{R}^{N_q \times C}$ for the query image and $F_{nc} \in \mathbb{R}^{K \times N_n \times C}$ for the normal images, where $N_q$ and $N_n$ represent the number of components in the query and normal images, respectively.

The feature maps $F_q$ and $F_n$ are subsequently passed through interpolation to obtain patch-level features, $P_q \in \mathbb{R}^{H_2 \times W_2 \times C}$ and $P_n \in \mathbb{R}^{K \times H_2 \times W_2 \times C}$. In parallel, textual descriptions of normal and anomalous semantics are processed by a text encoder, yielding textual features $T_n \in \mathbb{R}^C$ for normal semantics and $T_a \in \mathbb{R}^C$ for anomalous semantics. The patch-level features $P_q$, $P_n$, and textual features $T_n$ and $T_a$ are then input into the component-aware patch-matching module to generate a structural anomaly map.

At the same time, the component features of query image and normal images, $F_{qc}$ and $F_{nc}$, are passed through the graph-enhanced component modeling module to produce a logical anomaly map. Finally, UniVAD combines the structural and logical anomaly maps to generate the final unified anomaly detection result, as shown in Figure~\ref{fig:arch}.

In Sec 3.2 to 3.4, we provide a detailed description of contextual component clustering, component-aware patch matching, and graph-enhanced component modeling. Additionally, we provide pseudo-code descriptions of these modules in Appendix~\ref{app:A}.

\begin{figure}[t]
  \centering
   \includegraphics[width=\linewidth]{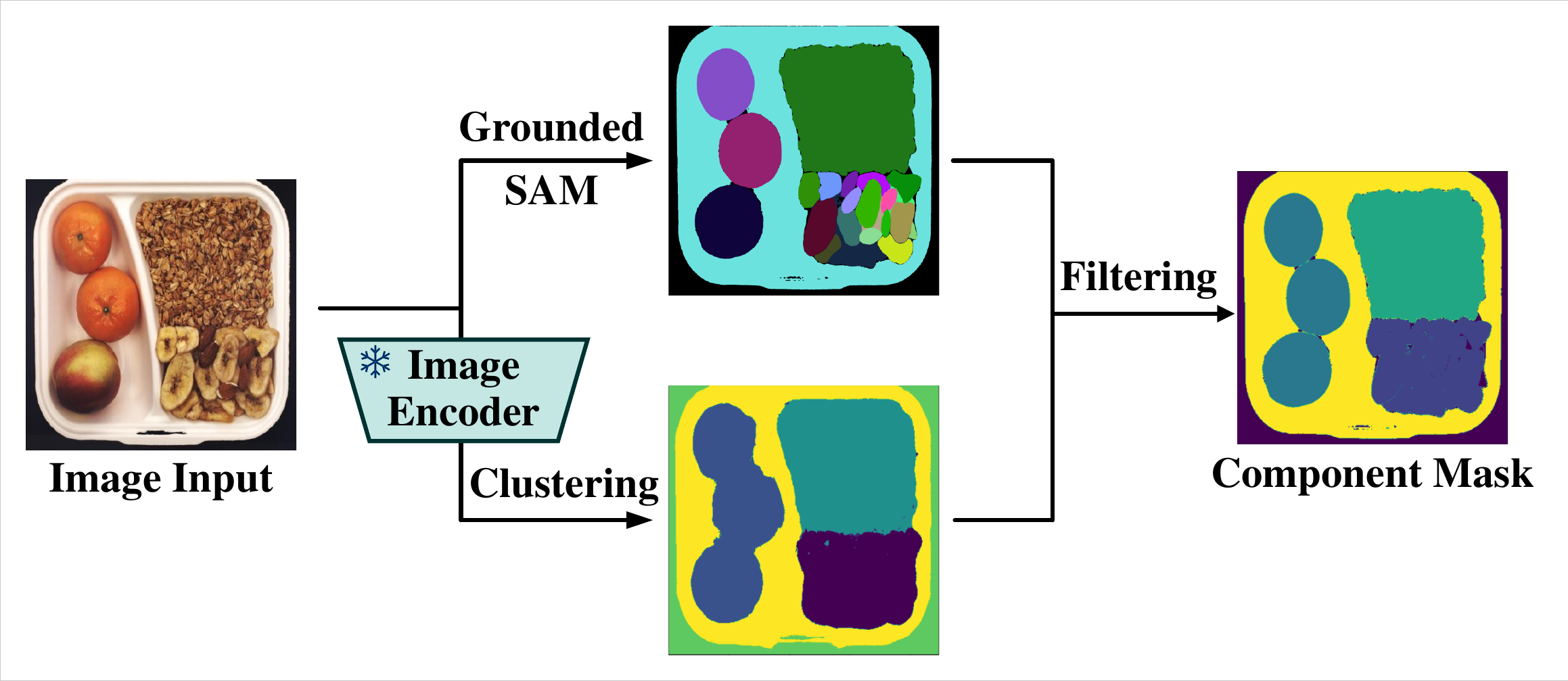}
   \caption{Architecture of the C\(^3\) module.}
   \label{fig:ccc}
\end{figure}

\subsection{Contextual Component Clustering}

To achieve accurate component segmentation with limited normal samples, we propose the Contextual Component Clustering (C\(^3\)) module, which combines visual foundation models with clustering techniques to enable precise component segmentation in few-shot setting.

Upon receiving an input image, the C\(^3\) module first uses the Recognize Anything Model~\cite{zhang2024recognize} to identify objects and generate content tags. Next, the Grounded SAM~\cite{ren2024grounded} method generates masks for all detected elements. However, SAM~\cite{kirillov2023segment} often faces challenges with segmentation granularity, producing masks that are either too fine or too coarse, which can be inconsistent across normal and query images. To address this, we refine and filter SAM’s output.

Specifically, after obtaining \( M \) initial component masks from Grounded SAM, represented as \( M_{\text{sam}} \in \mathbb{R}^{M \times H \times W} \), the C\(^3\) module evaluates the masks based on their quantity and area coverage. If only one mask is generated and it covers nearly the entire image (exceeding \( \gamma\% \) of the area), we infer that the image represents a textured surface (\eg, wood), and treat the entire image as a single component, outputting a mask covering the full image: \( M = \mathbf{1}_{H \times W} \).

If Grounded SAM generates a single mask that covers less than \( \gamma\% \) of the image area, we speculate that the image contains a single object, and we directly use the SAM-produced mask as the final mask: \( M = M_{\text{sam}} \).

In cases where Grounded SAM produces multiple masks, indicating the presence of multiple objects, we refine the SAM-generated masks using a clustering approach, as shown in Figure~\ref{fig:ccc}. Specifically, we first extract the feature map of the normal images \( F_n \in \mathbb{R}^{K \times H_1 \times W_1 \times C} \) using a pre-trained image encoder. Then, we apply the K-means clustering algorithm to the features in \( F_n \), clustering them into \( N \) groups and yielding \( N \) cluster centroids \( C \in \mathbb{R}^{N \times C} \). For each feature in the feature maps of both normal and query images, we compute its similarity to each cluster centroid, thereby generating \( N \) cluster masks for the image, denoted as \( M_{\text{cluster}} \in \mathbb{R}^{N \times H_1 \times W_1} \). After that, we filter the background masks by evaluating whether the values at the four corners of each mask are non-zero. This process yields \( N' \) valid masks, which are then resized to the original image dimensions to produce \( M_{\text{valid}} \in \mathbb{R}^{N' \times H \times W} \).

Next, for each mask \( M_{\text{sam}}^i \) in \( M_{\text{sam}} \), we compute the Intersection over Union (IoU) between \( M_{\text{sam}}^i \) and each mask in \( M_{\text{valid}} \), identifying the mask \( M_{\text{valid}}^j \) with the highest IoU. We then assign the label \( j \) to \( M_{\text{sam}}^i \):
\begin{equation}
  \text{Label}(M_{\text{sam}}^i) = \underset{j}{\arg\max} \ \text{IoU}(M_{\text{sam}}^i, M_{\text{valid}}^j).
\end{equation}

Finally, for each \( j \ (1 \leq j \leq N') \), the C\(^3\) module replaces the original mask \( M_{\text{valid}}^j \) with the union of the masks in \( M_{\text{sam}} \) that are assigned the label \( j \), resulting in the final mask set \( M \in \mathbb{R}^{N' \times H \times W} \):
\begin{equation}
  M^j = \bigcup_{\text{Label}(M_{\text{sam}}^i) = j} M_{\text{sam}}^i.
\end{equation}

This approach combines the precise segmentation capabilities of visual foundation models with the granularity control provided by clustering methods, enabling accurate component segmentation in few-shot settings.

Upon completing component segmentation, UniVAD proceeds to detect structural and logical anomalies in the image using the component-aware patch-matching module and the graph-enhanced component modeling module.

\subsection{Component-Aware Patch Matching}

Our Component-Aware Patch Matching (CAPM) method builds on patch feature matching, extending it by incorporating component constraints and image-text feature similarity comparisons to improve performance.

In the patch feature matching process, we first employ a network pretrained on large-scale datasets, such as ImageNet~\cite{ILSVRC15}, as an image encoder to extract the feature map of the query image, \( F_q \in \mathbb{R}^{H_1 \times W_1 \times C} \), and the feature map of the normal images, \( F_n \in \mathbb{R}^{K \times H_1 \times W_1 \times C} \). Next, both \( F_q \) and \( F_n \) are processed through interpolation to obtain patch features \( P_q \in \mathbb{R}^{H_2 \times W_2 \times C} \) and \( P_n \in \mathbb{R}^{K \times H_2 \times W_2 \times C} \), respectively. For each patch in the query image, we compute the cosine distance between \( P_q^i \) and all patch features in \( P_n \), using the minimum cosine distance as the patch-matching anomaly score for \( P_q^i \), as shown in Eq. (\ref{eq:com_patch}):
\begin{equation}
Score_{\text{pm}}(P_q^i) = \min(\text{distance}(P_q^i, P_n)).
\label{eq:com_patch}
\end{equation}

However, the standard patch feature matching method encounters limitations, such as its inability to distinguish between foreground and background regions, which may lead to false positives in background areas, and its failure to differentiate between different object components, which can mistakenly pair patches from other irrelevant regions with similar color or texture, resulting in missed detections. To address these issues, we leverage the component mask obtained from the \( C^3 \) module to perform feature matching within each component, effectively enhancing the accuracy of anomaly detection.

Specifically, after obtaining the patch features \( P_n \) of normal samples, we use the \( N' \) component masks derived from the \( C^3 \) module to create \( N' \) patch subsets \( P_{ni} \ (1 \leq i \leq N') \), where patches within each mask are allocated to their respective subsets:
\begin{equation}
P_{ni} = \{P_n^j \mid M_i^j = 1\}.
\end{equation}

We apply the same process to \( P_q \) to obtain \( P_{qi} \), and perform patch feature matching within each subset to calculate the component-aware anomaly score for each patch:
\begin{equation}
Score_{\text{aware}}(P_{qi}^j) = \min(\text{distance}(P_{qi}^j, P_{ni})).
\end{equation}

Additionally, we employ an image-text feature-matching approach to calculate an anomaly score for each patch. Specifically, we extract normal and anomalous text features, \( T_n \) and \( T_a \), using a pretrained text encoder that encodes the textual descriptions of normal and anomalous states. We then compute the cosine similarity between each patch feature of the query image and \( T_n \) and \( T_a \), obtaining the image-text anomaly score for each patch:
\begin{equation}
Score_{\text{vl}}(P_{qi}^j) = \text{softmax}(\text{sim}(P_{qi}^j, [T_n, T_a])),
\end{equation}
where \(\text{sim}(\cdot, \cdot)\) represents the cosine similarity calculation.

Finally, we sum the three anomaly scores for each patch to derive the structural anomaly score map:
\begin{equation}
Score_{\text{stru}} = \alpha  \, Score_{\text{pm}} + \beta  \, Score_{\text{aware}} + \gamma  \, Score_{\text{vl}},
\end{equation}
where \( \alpha \), $\beta$ and \( \gamma \) are hyperparameters, which are all set to $1/3$ in our experiments. 

\begin{table*}[]
\small
\centering
\caption{Comparison between UniVAD and existing methods under the 1-normal-shot setting, where image-level AUC and pixel-level AUC are used to evaluate the performance of image-level anomaly detection and pixel-level anomaly localization, respectively. The best results are highlighted in \textbf{bold}.}
\begin{tabular}{@{}ccccccccc@{}}
\toprule
Task                         & Dataset    & PatchCore & AnomalyGPT & WINCLIP & ComAD & UniAD & MedCLIP & \textbf{UniVAD (ours)} \\ \midrule
\multirow{9}{*}{\begin{tabular}[c]{@{}c@{}}Image-level\\ (AUC)\end{tabular}}
& MVTec-AD   & 84.0      & 94.1       & 93.1    & 57.3  & 70.3  & 75.2    & \textbf{97.8}          \\
& VisA       & 74.8      & 87.4       & 83.8    & 53.9  & 61.3  & 69.0    & \textbf{93.5}          \\
& MVTec LOCO & 62.0      & 60.4       & 58.0    & 62.2  & 50.9  & 54.9    & \textbf{71.0}          \\
& BrainMRI   & 73.2      & 73.1       & 55.4    & 33.3  & 50.0  & 69.7    & \textbf{80.2}          \\
& LiverCT    & 44.9      & 60.3       & 60.3    & 45.0  & 35.0  & 40.5    & \textbf{70.0}          \\
& RESC       & 56.3      & 82.4       & 72.9    & 73.5  & 53.5  & 66.9    & \textbf{85.5}          \\
& HIS        & 55.6      & 50.2       & 55.8    & 49.8  & 50.0  & 71.1    & \textbf{72.6}          \\
& ChestXray  & 66.4      & 68.5       & 70.2    & 50.1  & 60.6  & 71.4    & \textbf{72.2}          \\
& OCT17        & 59.9      & 77.5       & 79.7    & 57.6  & 44.4  & 64.6    & \textbf{82.1}          \\ \midrule
\multirow{6}{*}{\begin{tabular}[c]{@{}c@{}}Pixel-level\\ (AUC)\end{tabular}}
& MVTec-AD   & 89.9      & 95.3       & 95.2    & -     & 90.7  & 79.1    & \textbf{96.5}          \\
& VisA       & 93.4      & 96.2       & 96.2    & -     & 90.3  & 88.2    & \textbf{98.2}          \\
& MVTec LOCO & 69.8      & 70.3       & 58.8    & -     & 70.6  & 69.1    & \textbf{75.1}          \\
& BrainMRI   & 96.0      & 96.0       & 86.6    & -     & 93.6  & 91.7    & \textbf{96.8}          \\
& LiverCT    & 95.6      & 95.8       & 94.5    & -     & 88.5  & 93.8    & \textbf{96.3}          \\
& RESC       & 78.2      & 94.0       & 87.9    & -     & 80.7  & 91.5    & \textbf{94.9}          \\ \bottomrule
\end{tabular}
\label{tab:few-normal-shot}
\end{table*}

\subsection{Graph-Enhanced Component Modeling}

The CAPM module introduced earlier is primarily designed to detect structural anomalies with low-level semantics, where the anomalous content has never appeared in the normal samples. However, for higher-level semantic logical anomalies, the image content may exist in the normal samples but is combined incorrectly. Such anomalies are challenging to detect through patch feature matching alone, as they require a higher level of semantic understanding.

To address this problem, we design a Graph-Enhanced Component Modeling (GECM) module, which focuses on the holistic characteristics of each component, enabling it to detect the addition, omission, or misplacement of components. Specifically, after obtaining the component masks, the GECM module first employs a pretrained feature extractor to generate the feature map \( F_q \in \mathbb{R}^{H_1 \times W_1 \times C} \) for the query image and \( F_n \in \mathbb{R}^{K \times H_1 \times W_1 \times C} \) for the normal images. We then apply group average pooling to capture the deep features of each component from the query and normal samples, denoted as \( F_{qc} \in \mathbb{R}^{N_q \times C} \) and \( F_{nc} \in \mathbb{R}^{K \times N_n \times C} \). 

Next, we employ a Component Feature Aggregator~(CFA) module to further model each component's features. In the Component Feature Aggregator, we initially model each component feature as a node in a graph, and the cosine similarity between any two component features as the weight of the edge connecting these nodes, allowing us to compute the adjacency matrix for all nodes in the graph:

\begin{equation}
A = \begin{bmatrix}
S_{11} & S_{12}  & \cdots   & S_{1N}   \\
S_{21} & S_{22}  & \cdots   & S_{2N}  \\
\vdots & \vdots  & \ddots   & \vdots  \\
S_{N1} & S_{N2}  & \cdots\  & S_{NN}  \\
\end{bmatrix},
\end{equation}
where \textit{N} denotes the number of components~($N_q$ for the query image and $N_n$ for normal images), and \( S_{ij} \) represents the normalized similarity between nodes \( i \) and \( j \), defined as:
\begin{equation}
S_{i j} = \frac{S_{i j}^{\prime}}{\sum_{k=1}^N S_{i k}^{\prime}}, S_{ij}^{\prime} = \text{sim}(node_i,node_j).
\end{equation}

Subsequently, we leverage the adjacency matrix \( A \) to aggregate node information again via a graph attention operation, resulting in feature embeddings that more comprehensively represent the overall characteristics of each component. Specifically, these embeddings are expressed as \( E_q = G(A_q, F_{qc}) \) and \( E_n = G(A_n, F_{nc}) \), where \( G \) represents the graph attention operation~\cite{xiang2023agca}.

For each component feature embeddings \( E_q^i \) in \( E_q \), we compute its minimum cosine distance to the vectors in \( E_n \), which serves as the deep anomaly score for that component:
\begin{equation}
    Score_{\text{deep}}(E_q^i) = \min(\text{distance}(E_q^i, E_n)).
\end{equation}

In addition to deep features, geometric features such as component area, color, and position are also effective for detecting logical anomalies. Therefore, we compute these geometric features for each component in both query and normal samples, combining them into geometric feature vectors, represented as $ G_q \in \mathbb{R}^{N_q \times C_g} $ and \( G_n \in \mathbb{R}^{N_n \times C_g} \), and utilize the following formula to calculate the geometric anomaly score:
\begin{equation}
    Score_{\text{geo}}(G_q^i) = \min(\text{distance}(G_q^i, G_n)).
\end{equation}

Then, we combine \( Score_{deep} \) and \( Score_{geo} \) to obtain the logical anomaly score:
\begin{equation}
    Score_\text{logic} = \phi  \, Score_{\text{deep}} + \psi  \, Score_{\text{geo}}.
\end{equation}
where \( \phi \) and \( \psi \) are hyperparameters, and they are set to 0.5 in our experiments.

By combining \( \text{Score}_{\text{stru}} \) and \( \text{Score}_{\text{logic}} \), we derive the final anomaly score map:
\begin{equation}
Score_{\text{final}} = \delta \, Score_{\text{stru}} + \eta \, Score_{\text{logic}},
\end{equation}
where \( \delta \) and \( \eta \) are set to 0.5 by default.

\section{Experiments}

\subsection{Experimental Setups}
\textbf{Datasets.}  
We conduct extensive experiments on nine datasets spanning industrial, logical, and medical anomaly detection domains. For industrial anomaly detection, we use the widely recognized MVTec-AD~\cite{bergmann2019mvtec} and VisA~\cite{zou2022spot} datasets. For logical anomaly detection, we focus on the comprehensive MVTec LOCO~\cite{bergmann2022beyond} dataset. In the medical anomaly detection domain, following the recent BMAD benchmark, we select six datasets: BrainMRI~\cite{baid2021rsna}, liverCT~\cite{bilic2023liver, landman2015miccai}, RetinalOCT~\cite{hu2019automated}, ChestXray~\cite{wang2017chestx}, HIS~\cite{bejnordi2017diagnostic}, and OCT17~\cite{kermany2018identifying}. Since ChestXray~\cite{wang2017chestx}, HIS~\cite{bejnordi2017diagnostic}, and OCT17~\cite{kermany2018identifying} datasets do not provide pixel-level anomaly annotations, we evaluate only image-level anomaly detection performance on these three datasets. A detailed description of the nine datasets is provided in Appendix~\ref{app:B}.

\noindent\textbf{Competing Methods and Baselines.}  
In this study, we compare the performance of UniVAD with state-of-the-art methods from various domains, under two different settings, few-normal-shot and few-abnormal-shot. In the few-normal-shot setting, the model is not trained on the target dataset; instead, a few normal samples are provided only as reference during testing. Under this setting, we selected PatchCore~\cite{roth2022towards}, WinCLIP~\cite{jeong2023winclip}, AnomalyGPT~\cite{gu2024anomalygpt}, and UniAD~\cite{you2022unified} for industrial anomaly detection, ComAD~\cite{liu2023component} for logical anomaly detection, and MedCLIP~\cite{wang2022medclip} for medical anomaly detection. Few-abnormal-shot is a commonly used setting in medical anomaly detection, where testing is performed after training on a small number of normal and abnormal samples from the target dataset. To demonstrate the generality of UniVAD, we also compared it with existing methods in the few-abnormal-shot setting, selecting DRA~\cite{ding2022catching}, BGAD~\cite{yao2023explicit}, and MVFA~\cite{huang2024adapting} under this setting.

\noindent\textbf{Evaluation Protocols.}  
In alignment with established anomaly detection methodologies, we evaluate performance using the Area Under the Receiver Operating Characteristic Curve (AUC). Image-level AUC is used to assess anomaly detection performance, while pixel-level AUC is employed to evaluate anomaly localization performance.

\begin{table}[]
\centering
\small
\caption{Comparison between UniVAD and existing methods under the 4-abnormal-shot setting, where image-level AUC and pixel-level AUC are used to evaluate the performance of image-level anomaly detection and pixel-level anomaly localization, respectively. The experimental results in the table are cited from MVFA~\cite{huang2024adapting}, and the best results are highlighted in \textbf{bold}. }
\begin{tabular}{@{}cccccc@{}}
\toprule
Task                                                                         & Dataset   & DRA  & BGAD  & MVFA & \textbf{UniVAD} \\ \midrule
\multirow{6}{*}{\begin{tabular}[c]{@{}c@{}}Img-level\\ (AUC)\end{tabular}} 
& BrainMRI  & 80.6 & 83.6       & 92.4 & \textbf{94.1}          \\
& LiverCT   & 59.6 & 72.5       & 81.2 & \textbf{87.5}          \\
& RESC      & 90.9 & 86.2       & 96.2 & \textbf{97.3}          \\
& HIS       & 68.7 & -          & 82.7 & \textbf{85.7}          \\
& ChestXray & 75.8 & -          & 82.0 & \textbf{82.4}          \\
& OCT       & 99.0 & -          & 99.4 & \textbf{99.7}          \\ \midrule
\multirow{3}{*}{\begin{tabular}[c]{@{}c@{}}Px-level\\ (AUC)\end{tabular}} 
& BrainMRI  & 74.8 & 92.7       & 97.3 & \textbf{98.6}          \\
& LiverCT   & 71.8 & 98.9       & \textbf{99.7} & \textbf{99.7}          \\
& RESC      & 77.3 & 93.8       & \textbf{99.0} & \textbf{99.0}          \\ \bottomrule
\end{tabular}
\label{tab:few-abnormal-shot}
\end{table}

\noindent\textbf{Implementation Details.}
In few-normal-shot setting, we do not conduct any further training on anomaly detection datasets for UniVAD. We resize all images to a resolution of 448x448 pixels, and utilize two widely used vision encoders, CLIP-L/14@336px and DINOv2-G/14, as our image encoders, with their parameters frozen. For image-level anomaly scores, it is common to derive results from pixel-level outputs using a post-processing method. Depending on the distribution of detection data, popular approaches include using either the maximum or the mean of pixel-level results. In UniVAD, for datasets like HIS~\cite{bejnordi2017diagnostic}, where abnormal samples (\eg, cancer cell-stained slides) exhibit global differences compared to normal samples, we use the mean of the pixel-level results as the image-level anomaly score. For industrial anomaly detection, logical anomaly detection, and the remaining medical anomaly detection datasets, where abnormal regions occupy only a small portion of the image and the rest remains normal, we use the maximum of pixel-level results as the global anomaly score.

\begin{table}[]
\centering
\small
\caption{Comparison of different implementations of the $C^3$ module across multiple datasets. \textit{G-SAM Only} in table refers to the use of visual foundation models exclusively. The best performance results are highlighted in \textbf{bold}.}
\begin{tabular}{@{}cccc@{}}
\toprule
Dataset    & Cluster Only & G-SAM Only   & \textbf{$C^3$ Module} \\ \midrule
MVTec-AD   & (97.3, 96.1) & (97.5, 96.1) & \textbf{(97.8, 96.5)} \\
VisA       & (92.5, \textbf{98.0})   & (92.1, 97.7) & \textbf{(93.5, 98.0)}   \\
MVTec LOCO & (67.5, 70.9) & (67.8, 74.9)       & \textbf{(71.0, 75.1)} \\
BrainMRI   & (73.9, 96.7) & (74.5, 94.9) & \textbf{(80.2, 96.8)} \\
LiverCT    & (63.6, 96.1) & (63.7, \textbf{96.3}) & \textbf{(70.0, 96.3)}   \\
RESC       & (84.2, \textbf{94.9}) & (85.0, 94.6)   & \textbf{(85.5, 94.9)} \\ \bottomrule
\end{tabular}
\label{tab:abccc}
\end{table}

\subsection{Main Results}
\textbf{Few-normal-shot Setting.}
We conduct experiments using the same few-normal-shot setting as most existing few-shot anomaly detection methods~\cite{gu2024anomalygpt, jeong2023winclip, roth2022towards}, where the model is tested on objects it has never encountered during training, with a small number of normal samples provided as references during testing. Table~\ref{tab:few-normal-shot} presents a comparison of the performance of UniVAD and various domain-specific anomaly detection methods under the 1-normal-shot setting. It can be observed that UniVAD significantly outperforms existing domain-specific methods in both image-level and pixel-level results across different domains. Compared to state-of-the-art methods in each domain, our approach achieves an average improvement of 6.2\% in image-level AUC and 1.7\% in pixel-level AUC. The experimental results demonstrate the strong transferability of UniVAD.


\begin{table}[]
\centering
\small
\caption{Comparison of the clustering part in the $C^3$ module using image features extracted by different image encoders.}
\begin{tabular}{@{}cccc@{}}
\toprule
Dataset    & CLIP-ViT     & DINO         & DINOv2       \\ \midrule
MVTec-AD   & (97.7, 96.5) & (97.8, 96.5) & (97.9, 96.4) \\
VisA       & (93.5, 98.0) & (93.5, 98.0) & (93.2, 98.0) \\
MVTec LOCO & (71.1, 74.5) & (71.0, 75.1) & (70.9, 74.6) \\
RESC       & (85.2, 94.6) & (85.5, 94.9) & (85.2, 94.8) \\ \bottomrule
\end{tabular}
\label{tab:abbn}
\end{table}

\noindent\textbf{Few-abnormal-shot Setting.}
One of the key features of UniVAD is its robust generalization capability. Without requiring any training on domain-specific anomaly detection datasets, UniVAD demonstrates outstanding cross-domain anomaly detection performance by using only a minimal number of normal samples as a reference during the testing phase. On the other hand, for scenarios demanding high-precision detection on specific domain data, we provide a domain adaptation training method. This method allows UniVAD to be fine-tuned on domain-specific datasets to achieve optimal performance for particular tasks. Such fine-tuning requires only a small number of normal and anomalous samples from the target dataset, which is referred to as the few-abnormal-shot setting. Several popular medical anomaly detection methods~\cite{huang2024adapting, ding2022catching, yao2023explicit} employ the few-abnormal-shot setting for experimentation. We apply the same approach to UniVAD and present a comparison of its performance with these methods under the 4-abnormal-shot setting across six medical datasets in Table~\ref{tab:few-abnormal-shot}. Experimental results indicate that, under the few-abnormal-shot setting, UniVAD also outperforms existing approaches. Implementation details about UniVAD in the few-abnromal-shot setting can be found in Appendix~\ref{app:C}.

\subsection{Ablation Study}

We conduct extensive ablation studies on our proposed modules to demonstrate their effectiveness. Here, we primarily present the ablation results of our core module, with additional ablation study results provided in Appendix~\ref{app:D}.

\noindent\textbf{Contextual Component Clustering.}
Contextual Component Clustering ($C^3$) module id build upon visual foundation models such as RAM~\cite{zhang2024recognize}, Grounding DINO~\cite{liu2023grounding}, and SAM~\cite{kirillov2023segment}, as well as clustering techniques. It addresses the challenges of poor performance in few-shot settings commonly encountered in clustering methods and the difficulty of controlling segmentation granularity in SAM~\cite{kirillov2023segment}. In Table~\ref{tab:abccc}, we present a comparative analysis of performance results using only visual foundation models, only clustering methods, and the $C^3$ module, demonstrating the effectiveness of the $C^3$ module. We also provide a comparison of different image encoders utilized in clustering in Table~\ref{tab:abbn}. The performance differences are minimal, demonstrating the robustness of the $C^3$ module across various encoders.

\noindent\textbf{Component-Aware Patch Matching.}
The original patch feature matching method matches all patch features from the entire image, which can mistakenly pair patches from background or other irrelevant regions with similar color or texture, leading to decreased anomaly detection performance. In contrast, the CAPM module restricts the matching regions of patch features, ensuring that source and target patches originate from the same part. Table~\ref{tab:ab1} compares the performance of the original patch matching method and CAPM across multiple datasets, highlighting the effectiveness of CAPM in detecting structural anomalies.

\begin{table}[]
\centering
\small
\caption{Comparison between Patch Matching and CAPM methods on MVTec-AD, VisA, MVTec LOCO and BrainMRI datasets. \textit{Img-AUC} and \textit{Px-AUC} in table represent image-level AUC and pixel-level AUC. The best performance results are in \textbf{bold}.}
\begin{tabular}{@{}ccccc@{}}
\toprule
\multirow{2}{*}{Dataset} & \multicolumn{2}{c}{Patch Matching} & \multicolumn{2}{c}{CAPM} \\ \cmidrule(l){2-5} 
                         & Img-AUC        & Px-AUC       & Img-AUC   & Px-AUC  \\ \midrule
MVTec-AD                 & 96.9             & 96.3            & \textbf{97.8}        & \textbf{96.5}       \\
VisA                     & 91.8             & 97.8            & \textbf{93.4}        & \textbf{98.2}        \\
MVTec LOCO               & 62.0             & 69.8            & \textbf{64.1}        & \textbf{70.2}       \\
BrainMRI               & 79.1             & 96.4            & \textbf{80.2}        & \textbf{96.8}       \\
\bottomrule
\end{tabular}

\label{tab:ab1}
\end{table}

\noindent\textbf{Graph-Enhanced Component Modeling.}
Patch features with low-level semantics struggle to capture the overall characteristics of each component. Our GECM method, built on a graph neural network, models the interaction between each component’s geometric and deep features, enhancing the detection performance for logical anomalies with high-level semantics. Table~\ref{tab:ab2} compares the performance differences from using only CAPM to incrementally adding each module in GECM, demonstrating GECM’s notable effectiveness in detecting logical anomalies.

\begin{table}[]
\centering
\small
\caption{Comparison on the MVTec LOCO dataset under different settings. \textit{Geo feat.} represent components geometric features, \textit{Deep feat.} represents component deep features, and \textit{CFA} represents component feature aggregator. The best performance results are highlighted in \textbf{bold}.}
\begin{tabular}{@{}ccccc@{}}
\toprule
\multicolumn{3}{c}{Modules in GECM}                    & \multicolumn{2}{c}{MVTec LOCO} \\ \midrule
Geo feat. & Deep feat. & CFA                  & Image-AUC      & Pixel-AUC     \\ \midrule
                     &                       &                      & 64.1             & 70.2          \\
$\checkmark$         &                       &                      & 66.6             & 73.2          \\
$\checkmark$         & $\checkmark$          & \multicolumn{1}{l}{} & 69.4             & 74.8          \\
$\checkmark$         & $\checkmark$          & $\checkmark$                    & \textbf{71.0}             & \textbf{75.1}          \\ \bottomrule
\end{tabular}
\label{tab:ab2}
\end{table}

\begin{figure}[t]
  \centering
   \includegraphics[width=\linewidth]{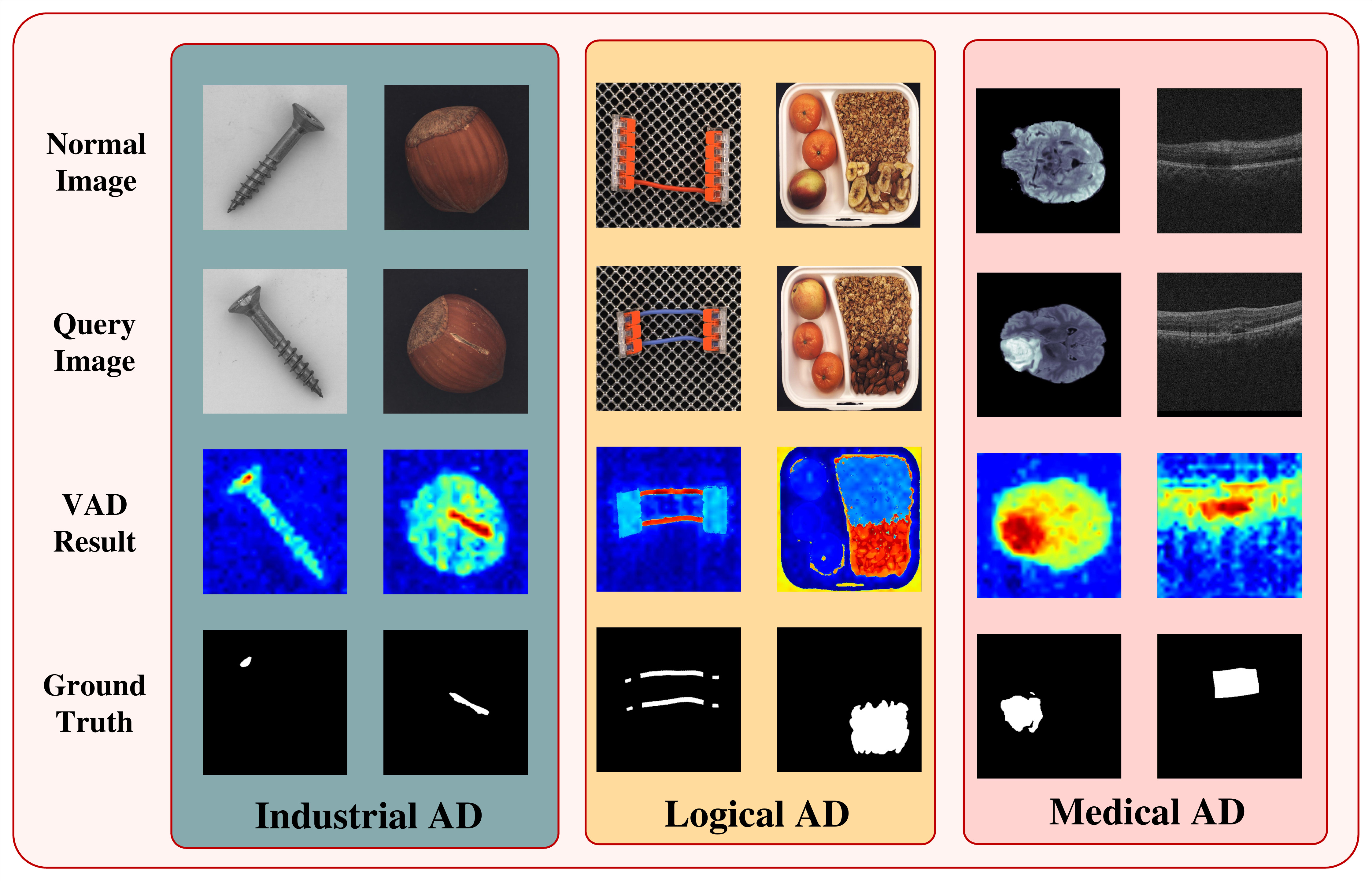}
   \caption{Visualization result of UniVAD on datasets across diverse domains. UniVAD demonstrates a strong transferability by accurately segmenting previously unseen samples with only a single normal sample provided as reference.}
   \label{fig:visual}
\end{figure}

\subsection{Visualization Results}
Figure~\ref{fig:visual} illustrates the visual anomaly detection results of UniVAD across various datasets in industrial anomaly detection, logic anomaly detection, and medical anomaly detection. With only a single normal sample as a reference, UniVAD accurately detects anomalies in previously unseen items across diverse domains, demonstrating the method’s strong transferability and practical applicability.
\section{Limitation}
UniVAD currently relies on visual foundation models such as RAM~\cite{zhang2024recognize}, Grounding DINO~\cite{liu2023grounding}, and SAM~\cite{kirillov2023segment} for component segmentation. The computational latency during inference somewhat limits its applicability in real-time scenarios. However, the key strength of UniVAD lies in its remarkable generalization capability. Even when provided with only a small number of normal samples as references, UniVAD is able to accurately detect anomalies across diverse domains, thereby advancing the standardization of anomaly detection research.
\section{Conclusion}
In this paper, we propose UniVAD, a novel, training-free, unified few-shot visual anomaly detection method capable of detecting anomalies across various domains, including industrial, logical, and medical fields, using a unified model. By leveraging a contextual component clustering module for precise segmentation, along with component-aware patch matching and graph-enhanced component modeling for multi-level anomaly detection, UniVAD achieves superior performance without the need for domain-specific models or extensive training data. Experimental results across multiple datasets demonstrate that UniVAD outperforms existing domain-specific approaches, offering a more flexible and scalable solution for visual anomaly detection tasks and contributing to the standardization of research in the field of visual anomaly detection.

{
    \small
    \bibliographystyle{ieeenat_fullname}
    \bibliography{main}
}

\clearpage
\setcounter{page}{1}
\maketitlesupplementary

\appendix

\renewcommand\thesection{\Alph{section}}
\setcounter{section}{0}

\section{Pseudocode Descriptions}
\label{app:A}
In this section, we present Pytorch-style pseudocode for the three proposed modules: Contextual Component Clustering ($C^3$), Component-Aware Patch Matching (CAPM), and Graph-Enhanced Component Modeling (GECM). These pseudocode representations aim to provide readers with a clearer and more structured understanding of the implementation details for each module. Specifically, the descriptions cover key processes such as mask filtering in the $C^3$ module, patch matching in the CAPM module, and component feature aggregation in the GECM module. Algorithm~\ref{algo:code_ccc}, Algorithm~\ref{algo:code_capm}, and Algorithm~\ref{algo:code_gecm} illustrate the pseudocode for these three modules, respectively.

In the GECM module, a graph structure is employed to enhance the interaction modeling of features for each component in the image. Specifically, the features of individual components, obtained through group average pooling, are treated as nodes in the graph. The edges between these nodes are weighted based on the cosine similarity between their corresponding features, capturing the relational structure within the image. 

To further model these relationships, we leverage a training-free graph attention mechanism, which facilitates the exchange of information among the graph's nodes. In this approach, the feature of each node simultaneously serves as the Query (Q), Key (K), and Value (V). The attention mechanism is computed using the following formula:

\begin{equation}
    Attention(Q, K, V) = norm\left(\frac{Q \cdot K^T}{\sqrt{d_k}}\right)V,
\end{equation}
where \( norm(\cdot) \) represents a normalization operation that ensures each row of the attention matrix sums to 1. A widely used normalization technique is the \( softmax(\cdot) \) function, which applies an exponential scaling to emphasize relative importance among nodes. This approach allows for effective and efficient feature interaction modeling, leveraging the inherent structure of the graph without requiring additional training.


\begin{algorithm}
\caption{Pseudocode of $C^3$ in a PyTorch style.}
\label{algo:code_ccc}
\begin{lstlisting}[language=python]
def C_3(image):

    # generate M_sam
    tags = RAM(image)
    M_sam = Grounded_SAM(image, tags)
    M, H, W = M_sam.shape
    
    # when M_sam contains only one mask
    if len(M_sam) == 1:
        area_ratio = M_sam[0].sum() / (H * W)
        if area_ratio > gamma:
            return ones_like(image)
        else:
            return M_sam
    
    # generate M_cluster
    image_features = image_encoder(image)
    cluster_centers = Cluster(image_features)
    # calculate  feature similarity
    M_cluster = sim(
        image_features, cluster_centers
    ).argmax()
    
    # combine masks
    for mask in M_sam:
        label[mask] = iou(mask, M_cluser).argmax()
        M_final[label[mask]].add(mask)

    return M_final
    
\end{lstlisting}
\end{algorithm}

\begin{algorithm}
\caption{Pseudocode of CAPM in a Pytorch style.}
\label{algo:code_capm}
\begin{lstlisting}[language=python]
def CAPM(image, normal_image, text_features):
    # get patch features
    query_feat = image_encoder(image)
    normal_feat = image_encoder(normal_image)
    query_patches = Interpolate(
        query_feat, size=image_size/patch_size
    )
    normal_patches = Interpolate(
        normal_feat, size=image_size/patch_size
    )

    # calculate score_pm
    distance_matrix = cos_distance(
        query_patches, normal_patches
    )
    score_pm = distance_matrix.min()

    # calculate score_aware and score_vl
    query_masks = C_3(image)
    normal_masks = C_3(normal_image)
    
    for mask, normal_mask in zip(
        query_masks, normal_masks
    ):
        distance_matrix = cos_distance(
            query_patches[mask], 
            normal_patches[normal_mask]
        )
        score_capm[mask] = distance_matrix.min()
    
        score_vl[mask] = cos_distance(
            query_patches[mask], text_features
        )

    return (score_pm + score_aware + score_vl) / 3
\end{lstlisting}
\end{algorithm}


\begin{algorithm}
\caption{Pseudocode of GECM in a Pytorch style.}
\label{algo:code_gecm}
\begin{lstlisting}[language=python]
def GECM(image, normal_image):

    # extract features
    query_feat = image_encoder(image)
    normal_feat = image_encoder(normal_image)
    query_masks = C_3(image)
    normal_masks = C_3(normal_image)
    
    query_com_feat = {"deep": [], "geo": []}
    normal_com_feat = {"deep": [], "geo": []}

    # CFA: Component Feature Aggregator
    # GAP: Group Average Pooling
    query_com_feat["deep"] = CFA(
        GAP(query_feat, query_masks)
    )
    normal_com_feat["deep"] = CFA(
        GAP(normal_feat, normal_masks)
    )

    query_com_feat["geo"] = geo_encoder(
        query_feat, query_masks
    )
    normal_com_feat["geo"] = geo_encoder(
        normal_feat, normal_masks
    )

    # calculate scores
    for mask in query_masks:
        dis_deep[mask] = cos_distance(
            query_com_feat["deep"][mask], 
            normal_com_feat["deep"][normal_masks]
        )
    
        dis_geo[mask] = cos_distance(
            query_com_feat["geo"][mask], 
            normal_com_feat["geo"][normal_masks]
        )
    
        score_deep[mask] = dis_matrix_deep.min()
        score_geo[mask] = dis_matrix_geo.min()

    
    return (score_deep + score_geo) / 2
    
\end{lstlisting}
\end{algorithm}

\section{Dataset Details}
\label{app:B}

We conduct extensive experiments on UniVAD using nine different datasets from the fields of industrial anomaly detection, logical anomaly detection, and medical anomaly detection, respectively. The following provides a detailed description of each dataset:

\noindent \textbf{MVTec-AD~\cite{bergmann2019mvtec}} is one of the most popular datasets for industrial anomaly detection tasks, consisting of 5,354 images across 15 different object categories. This includes 4,096 normal images and 1,258 anomalous images, with resolutions ranging from 700$\times$700 to 1,024$\times$1,024. The dataset covers various common industrial products, such as wooden boards, leather, metal components, and pills.

\noindent \textbf{VisA~\cite{zou2022spot}} is a relatively recent and widely used industrial anomaly detection dataset, containing 10,821 images across 12 object categories. It includes 2,162 anomalous images, with resolutions around 1,500$\times$1,000 pixels.

\noindent \textbf{MVTec LOCO~\cite{bergmann2022beyond}} is a dataset for logical anomaly detection. It is currently the largest logical anomaly detection dataset, containing 2,076 normal images and 1,568 anomalous images across 5 object categories. The anomalies in the dataset include both structural anomalies such as damage or defects, and logical anomalies such as the addition, omission, or incorrect combination of elements. The dataset provides pixel-level annotations of the anomalies.

\noindent \textbf{BrainMRI~\cite{bao2024bmad}} dataset is based on the BraTS2021~\cite{baid2021rsna} dataset, one of the latest large-scale brain tumor segmentation datasets. It contains complete 3D brain volume images. The BrainMRI dataset consists of 2D slices derived from BraTS2021, with each slice image measuring 240$\times$240 pixels. The training set includes 7,500 normal samples, and the test set contains 3,715 samples, both normal and anomalous, with pixel-level anomaly annotations.

\noindent \textbf{LiverCT~\cite{bao2024bmad}} dataset is constructed from the BTCV~\cite{landman2015miccai} and LiTS~\cite{bilic2023liver} datasets. It contains 50 normal abdominal 3D CT scans from BTCV and 131 abdominal 3D CT scans, both normal and anomalous, from LiTS. The Hounsfield Unit (HU) values of the 3D scans from both datasets are converted to grayscale using the abdominal window and then cropped into 2D slices. The dataset includes 1,452 normal 2D slices for training and 1,493 2D slices, both normal and anomalous, for testing, with a resolution of 512$\times$512 and pixel-level anomaly annotations.

\noindent \textbf{RESC~\cite{hu2019automated}}: The Retinal Edema Segmentation Challenge (RESC) dataset is a retinal OCT dataset containing 4,297 normal images for training and 1,805 test images, both normal and anomalous. The image resolution is 512$\times$1,024, and the dataset provides pixel-level anomaly annotations.

\noindent \textbf{OCT17~\cite{kermany2018identifying}} is another retinal OCT dataset, which includes 26,315 normal training images and 968 test images, both normal and anomalous, with a resolution of 512$\times$496. The dataset only provides image-level anomaly annotations.

\noindent \textbf{ChestXray~\cite{wang2017chestx}} dataset is a commonly used X-ray dataset for detecting pulmonary abnormalities. It contains 8,000 normal images for training and 17,194 test images, both normal and anomalous, with a resolution of 1,024$\times$1,024. The dataset provides image-level anomaly annotations.

\noindent \textbf{HIS~\cite{bao2024bmad}} dataset is cropped from the Camelyon16~\cite{bejnordi2017diagnostic} dataset, which includes 400 whole-slide images of lymph node biopsies stained with hematoxylin and eosin from breast cancer patients. The HIS dataset includes 6,091 normal image patches and 997 anomalous image patches, with a resolution of 256$\times$256 pixels. The dataset provides image-level anomaly annotations.

\begin{figure*}[t]
  \centering
   \includegraphics[width=\textwidth]{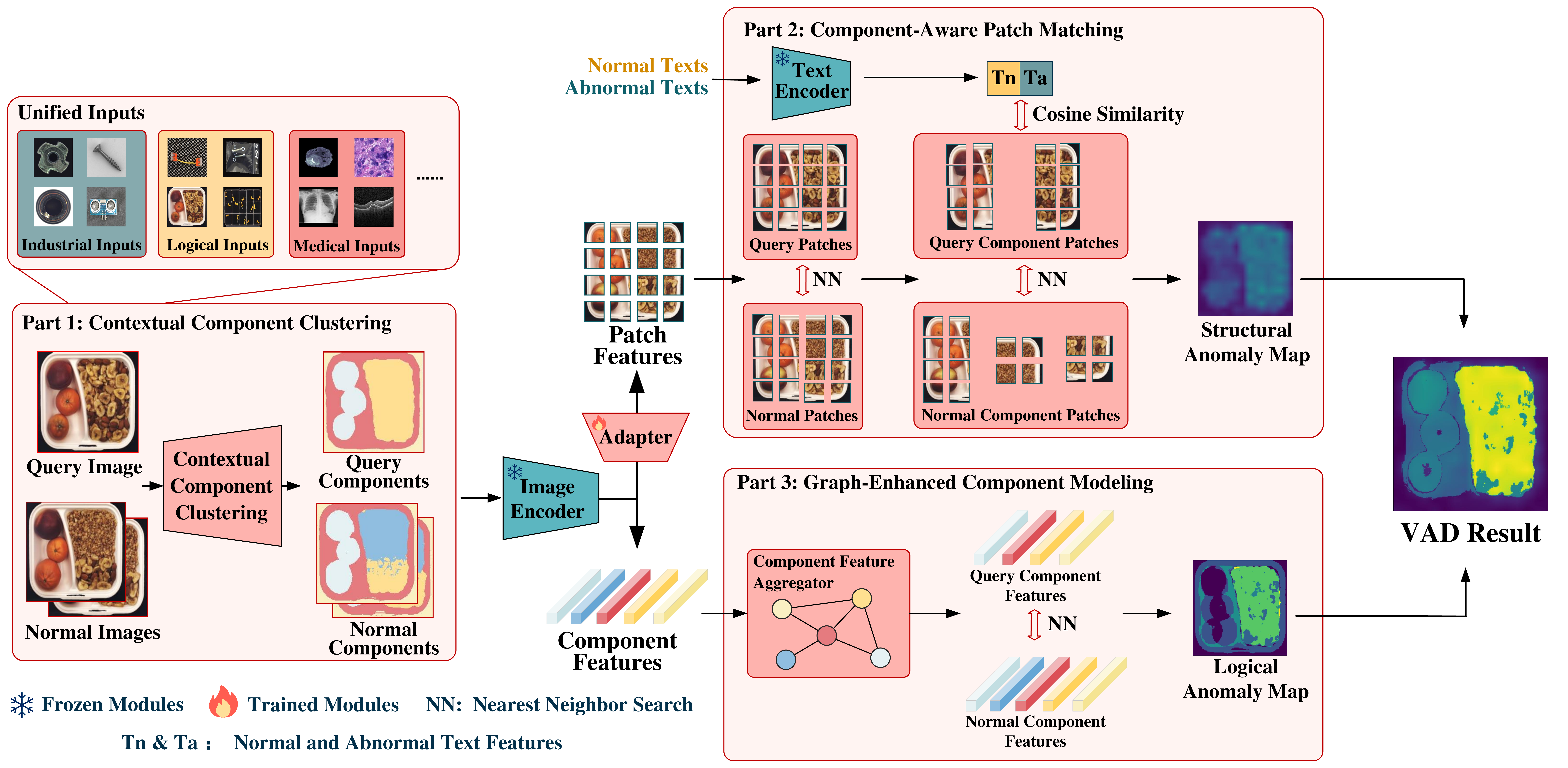}
   \caption{The overall architecture of UniVAD with a trainable adapter under few-abnormal-shot seeting. }
   \label{fig:arch_adapter}
\end{figure*}

\section{Few-Abnormal-Shot Setting}
\label{app:C}
One of the key features of UniVAD is its robust generalization capability. Without requiring any training on domain-specific anomaly detection datasets, UniVAD demonstrates outstanding cross-domain anomaly detection performance by using only a minimal number of normal samples as a reference during the testing phase.

On the other hand, for scenarios demanding high-precision detection on specific domain data, we provide a domain adaptation training method. This method allows UniVAD to be fine-tuned on domain-specific datasets to achieve optimal performance for particular tasks. Such fine-tuning requires only a small number of normal and anomalous samples from the target dataset, which is referred to as the few-abnormal-shot setting.

In Section 4, in addition to the default few-normal-shot setting, we conduct experiments using the few-abnormal-shot setting and compare its performance with other methods under the same configuration. The experimental results significantly outperform existing approaches, demonstrating that UniVAD combines strong generalization capabilities with exceptional accuracy for domain-specific tasks.

Training UniVAD for domain adaptation under the few-abnormal-shot setting requires only minimal modifications to the original model: adding an adapter after the image encoder, as illustrated in Figure~\ref{fig:arch_adapter}.

Regarding the adapter's structure, we adopt bottleneck architecture, which is commonly used in computer vision and natural language processing. The specific structure, shown in Algorithm~\ref{alg:adapter}, consists of two linear layers, one ReLU activation layer, and one SiLU activation layer.

\begin{algorithm}[h]
\caption{Adapter Module}
\label{alg:adapter}

\textbf{Input:} vector $\mathbf{x} $

\textbf{Output:} vector $\mathbf{y} $
\begin{algorithmic}[1]
\STATE $\mathbf{h}_1 = \text{ReLU}(\mathbf{W}_1 \mathbf{x} + \mathbf{b}_1)$ 
\STATE $\mathbf{y} = \text{SiLU}(\mathbf{W}_2 \mathbf{h}_1 + \mathbf{b}_2)$ 
\end{algorithmic}
\end{algorithm}

\begin{table*}[]
\centering
\begin{tabular}{@{}ccccc@{}}
\toprule
\multirow{2}{*}{Module} & \multicolumn{2}{c}{MVTec LOCO Logical}                                    & \multicolumn{2}{c}{MVTec LOCO Structural}                                 \\ \cmidrule(l){2-5} 
                        & \multicolumn{1}{l}{Image-level AUC} & \multicolumn{1}{l}{Pixel-level AUC} & \multicolumn{1}{l}{Image-level AUC} & \multicolumn{1}{l}{Pixel-level AUC} \\ \midrule
CAPM Only               & 59.9                                & 66.9                                & 82.2                                & 93.0                                \\
GECM Only               & 65.0                                & 71.2                                & 56.9                                & 82.6                                \\
\textbf{UniVAD}         & \textbf{65.9}                       & \textbf{72.4}                       & \textbf{82.6}                       & \textbf{93.5}                       \\ \bottomrule
\end{tabular}
\caption{Ablation studies of CAPM and GECM modules on MVTec LOCO datasets. The best performance results are in \textbf{bold}.}
\label{tab:ab_stru_logi}
\end{table*}

\begin{figure*}[t]
  \centering
   \includegraphics[width=0.8\textwidth]{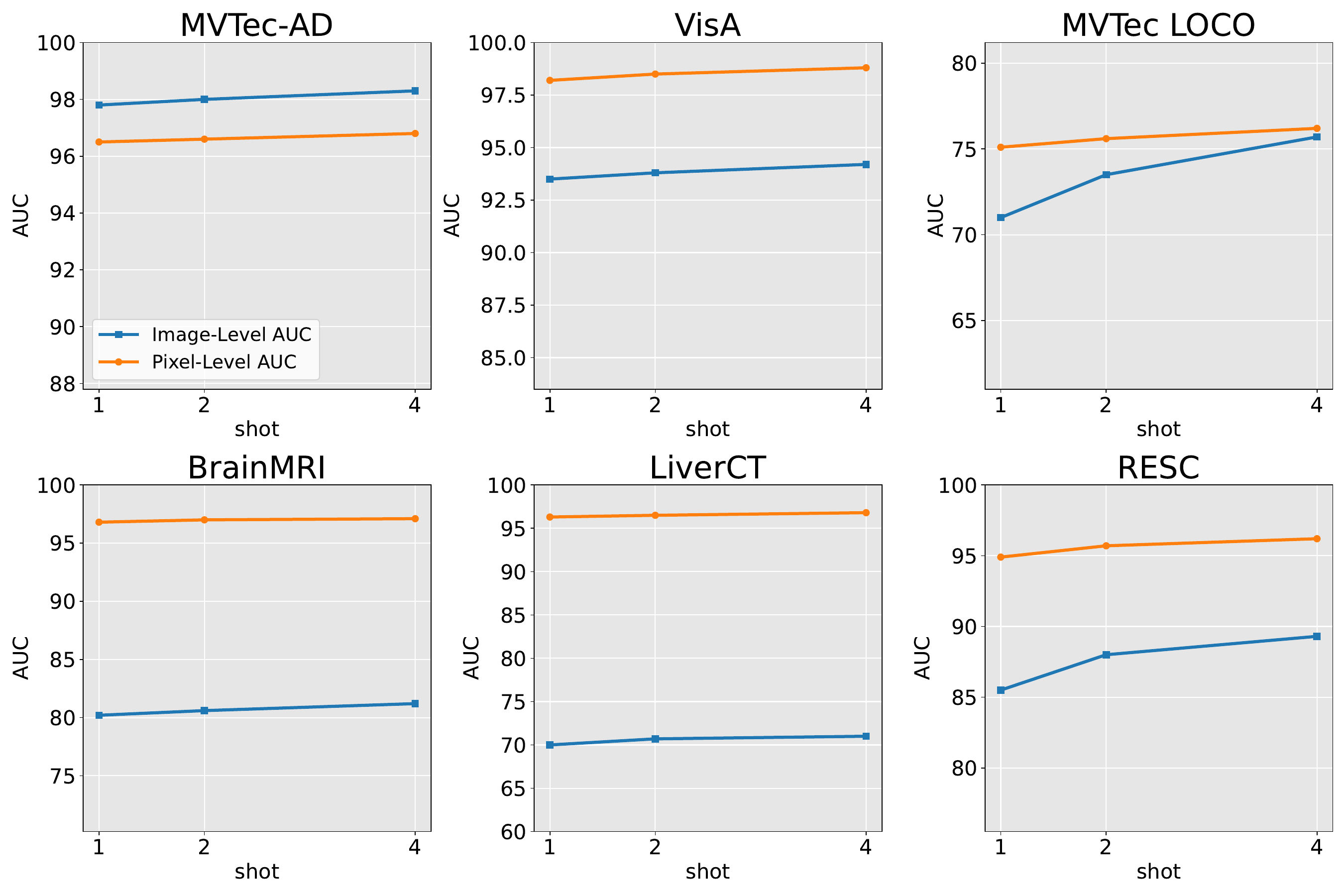}
   \caption{Experimental results of UniVAD under 1-normal-shot, 2-normal-shot, and 4-normal-shot settings.}
   \label{fig:shot}
\end{figure*}

\section{More ablation study results}
\label{app:D}
In this section, we present a more detailed ablation study of the various components of our proposed method. This includes an analysis of the hierarchical levels of features extracted by the image encoder, the geometric features utilized in the GECM module, the distance metrics employed in score computation, the number of normal samples considered, and other relevant factors. Below are the detailed analyses for each aspect.

\subsection{Structural Anomalies and Logical Anomalies}

Each category in the MVTec LOCO dataset contains both structural and logical anomalies. UniVAD's CAPM module and GECM module are particularly adept at detecting one type of anomaly each. In Table~\ref{tab:ab_stru_logi}, we compare the detection performance of the CAPM and GECM modules on the MVTec LOCO dataset, highlighting the complementary collaboration between the two modules.

\subsection{Normal Samples}
UniVAD can perform anomaly detection and localization across various domains with only a single normal sample as a reference. To further evaluate its performance, we conducted experiments under settings where multiple normal samples were provided as references. Specifically, experiments were performed with 1, 2, and 4 normal samples. The experimental results under different numbers of normal samples are presented in Fig.~\ref{fig:shot}. The results demonstrate that the anomaly detection performance improves progressively as the number of normal samples increases.

\begin{table*}[]
\centering

\begin{tabular}{@{}ccccccc@{}}
\toprule
Task                                                                         & Dataset    & Layer 6 Only & Layer 12 Only      & Layer 18 Only      & Layer 24 Only & \textbf{UniVAD} \\ \midrule
\multirow{6}{*}{\begin{tabular}[c]{@{}c@{}}Image-level\\ (AUC)\end{tabular}} & MVTec-AD   & 96.5    & 96.9          & 97.1          & 96.5     & \textbf{97.8}   \\
                                                                             & VisA       & 91.1    & 90.3          & 90.3          & 90.5     & \textbf{93.5}   \\
                                                                             & MVTec LOCO & 70.4    & 68.9          & 68.7          & 66.7     & \textbf{71.0}   \\
                                                                             & BrainMRI   & 68.9    & 66.2          & 73.9          & 68.8     & \textbf{80.2}   \\
                                                                             & LiverCT    & 59.0    & 63.4          & 64.4          & 65.5     & \textbf{70.0}   \\
                                                                             & RESC       & 78.2    & 84.1          & 84.1          & 83.5     & \textbf{85.5}   \\ \midrule
\multirow{6}{*}{\begin{tabular}[c]{@{}c@{}}Pixel-level\\ (AUC)\end{tabular}} & MVTec-AD   & 96.3    & \textbf{96.8} & 96.5          & 95.7     & 96.5            \\
                                                                             & VisA       & 95.6    & 97.9          & 97.9          & 97.3     & \textbf{98.2}   \\
                                                                             & MVTec LOCO & 74.7    & 74.6          & \textbf{75.5} & 73.2     & 75.1            \\
                                                                             & BrainMRI   & 95.3    & 95.8          & \textbf{96.9} & 95.4     & 96.8            \\
                                                                             & LiverCT    & 96.0    & 94.2          & 94.9          & 96.0     & \textbf{96.3}   \\
                                                                             & RESC       & 91.5    & 94.3          & 94.6          & 94.0     & \textbf{94.9}   \\ \bottomrule
\end{tabular}
\caption{Ablation studies of multi-level feature utilization across different datasets. The best performance results are in \textbf{bold}.}
\label{tab:ab-level}
\end{table*}

\subsection{Multi-level Features}
Numerous studies have demonstrated that image features extracted from different intermediate layers of an image encoder exhibit distinct characteristics. Shallow-layer features predominantly capture basic graphical properties such as colors and edges, while deep-layer features encapsulate more complex and abstract semantic information, including structures and textures. In UniVAD, four hierarchical feature maps are extracted from the input image, corresponding to layers 6, 12, 18, and 24 of the CLIP-ViT image encoder. The utilization of multi-level features enhances the representation capacity for features at varying levels of abstraction within the image, thereby improving anomaly detection performance. Table~\ref{tab:ab-level} compares the anomaly detection performance when using features from a single layer versus employing multi-level features.

\subsection{Geometric Features}
In the GECM module, in addition to leveraging the deep features of each component, we also incorporate the geometric features of each component to assess logical anomalies. In UniVAD, three geometric features, \ie area, color, and position, are extracted for each component and concatenated into a single vector. Table~\ref{tab:ab_geo} presents a comparison of anomaly detection performance across different datasets when only subsets of the geometric features are utilized. The results demonstrate that each geometric feature contributes significantly to the overall performance.

\subsection{Image Resolution}
\begin{figure*}[t]
  \centering
   \includegraphics[width=0.85\textwidth]{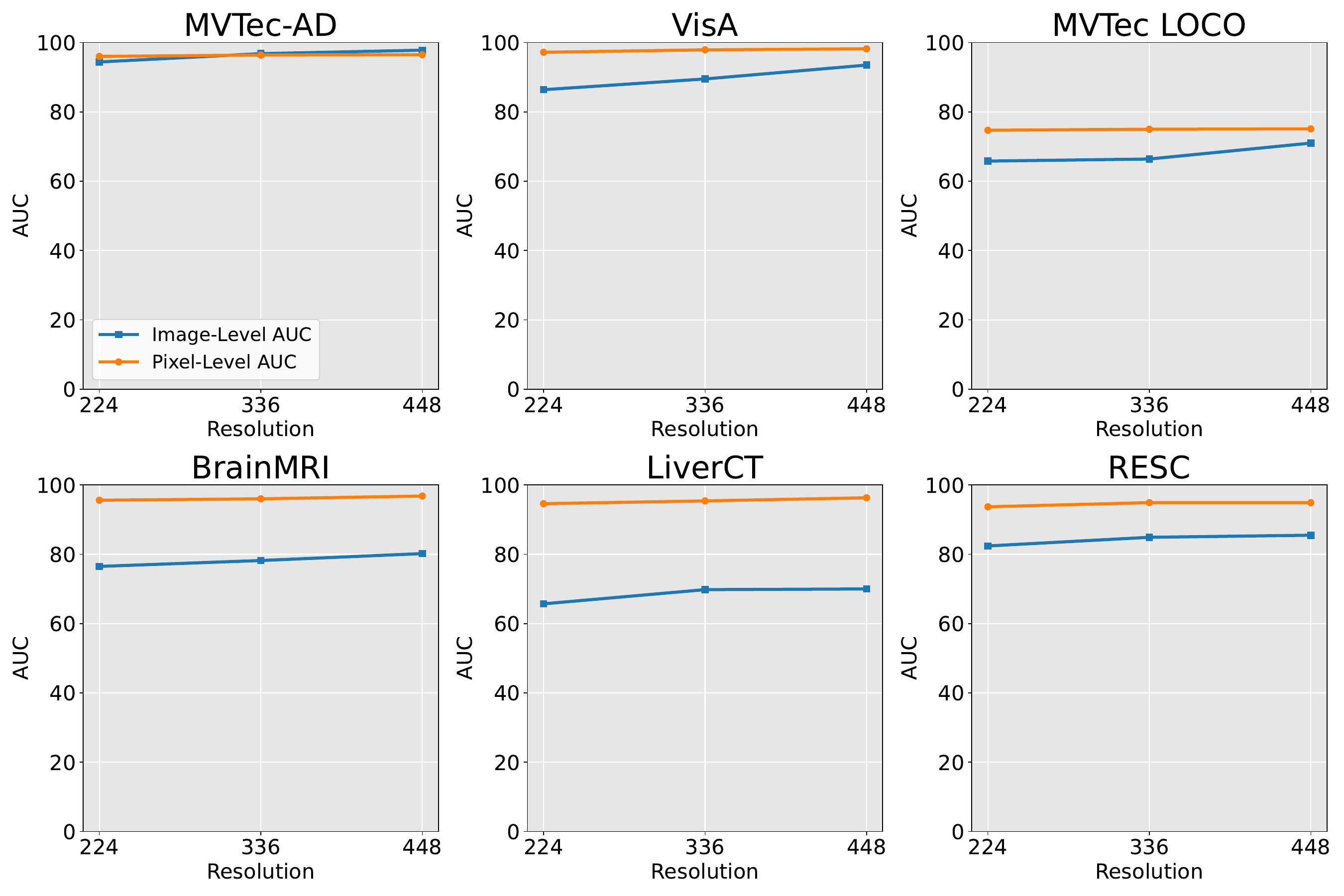}
   \caption{Experimental results of UniVAD at different resolutions.}
   \label{fig:resolution}
\end{figure*}

In the experiments presented in the main text, we adopt a resolution of 448$\times$448 to remain consistent with existing mainstream anomaly detection methods. To evaluate the performance of the method in scenarios with limited computational resources, we also test UniVAD at resolutions of 224$\times$224 and 336$\times$336. The experimental results shown in Fig.~\ref{fig:resolution} demonstrate that, although UniVAD experiences a slight performance drop under lower-resolution settings, it still achieves satisfactory results.

\subsection{Distance Calculation Method}
In the CAPM and GECM modules, UniVAD calculates the distances between image patch features and component features, respectively. By default, we use cosine distance for these calculations, as described in the paper. Additionally, we evaluated UniVAD's performance using L1 distance and L2 distance. The comparative experimental results are presented in Table~\ref{tab:ab_dis}. The results indicate that cosine distance achieves the best detection performance.

\subsection{Clustering Method}
In the $C^3$ module, we generate $M_{cluster}$ by clustering image features to filter and control the granularity of $M_{sam}$ produced by Grounded SAM. The clustering method used in the paper is KMeans. To investigate the impact of different clustering methods on performance, we also compared Meanshift, DBSCAN, and Spectral Clustering. The comparative results, summarized in Table~\ref{tab:ab_cluster}, show that anomaly detection performance is highest when using KMeans or Spectral Clustering, while Meanshift and DBSCAN yield slightly inferior results.

\begin{table*}[]
\centering
\begin{tabular}{@{}cccccc@{}}
\toprule
Task                                                                         & Dataset    & w/o Geo Feat.  & Area Only     & Area + Color  & \textbf{Area + Color + Position\ } \\ \midrule
\multirow{6}{*}{\begin{tabular}[c]{@{}c@{}}Image-level\\ (AUC)\end{tabular}} & MVTec-AD   & 97.4          & 97.3          & 97.6          & \textbf{97.8}                    \\
                                                                             & VisA       & 92.0          & 92.0          & 92.7          & \textbf{93.5}                    \\
                                                                             & MVTec LOCO & 69.8          & 70.6          & 70.8          & \textbf{71.0}                    \\
                                                                             & BrainMRI   & 79.7          & 79.4          & 79.6          & \textbf{80.2}                    \\
                                                                             & LiverCT    & 66.4          & 67.9          & 68.9          & \textbf{70.0}                    \\
                                                                             & RESC       & 83.9          & 84.2          & 84.9          & \textbf{85.5}                    \\ \midrule
\multirow{6}{*}{\begin{tabular}[c]{@{}c@{}}Pixel-level\\ (AUC)\end{tabular}} & MVTec-AD   & \textbf{96.6}          & \textbf{96.6} & 96.5          & 96.5                             \\
                                                                             & VisA       & 98.1          & 98.1          & 98.0          & \textbf{98.2}                    \\
                                                                             & MVTec LOCO & 72.4          & 72.1          & 74.9          & \textbf{75.1}                    \\
                                                                             & BrainMRI   & \textbf{96.8} & \textbf{96.8} & \textbf{96.8} & \textbf{96.8}                    \\
                                                                             & LiverCT    & 94.8          & 95.7          & 96.2          & \textbf{96.3}                    \\
                                                                             & RESC       & 94.4          & 94.4          & 94.7          & \textbf{94.9}                    \\ \bottomrule
\end{tabular}
\caption{Ablation studies of geometric features across different datasets, Geo feat. represent components geometric features. The best performance results are in \textbf{bold}.}
\label{tab:ab_geo}
\end{table*}

\begin{table*}[]
\centering
\begin{tabular}{@{}ccccc@{}}
\toprule
Task                                                                         & Dataset    & L1 distance & L2 distance & \textbf{Cosine distance\ } \\ \midrule
\multirow{6}{*}{\begin{tabular}[c]{@{}c@{}}Image-level\\ (AUC)\end{tabular}} & MVTec-AD   & 94.8        & 93.8        & \textbf{97.8}            \\
                                                                             & VisA       & 87.6        & 87.5        & \textbf{93.5}            \\
                                                                             & MVTec LOCO & 66.4        & 62.1        & \textbf{71.0}                     \\
                                                                             & BrainMRI   & 59.3        & 60.2        & \textbf{80.2}            \\
                                                                             & LiverCT    & 54.2        & 53.3        & \textbf{70.0}            \\
                                                                             & RESC       & 83.4        & 82.5        & \textbf{85.5}            \\ \midrule
\multirow{6}{*}{\begin{tabular}[c]{@{}c@{}}Pixel-level\\ (AUC)\end{tabular}} & MVTec-AD   & 95.3        & 95.7        & \textbf{96.5}            \\
                                                                             & VisA       & 96.7        & 97.1        & \textbf{98.2}            \\
                                                                             & MVTec LOCO & 75.1        & 69.7        & \textbf{75.1}            \\
                                                                             & BrainMRI   & 95.6        & 95.4        & \textbf{96.8}            \\
                                                                             & LiverCT    & 95.9        & 96.0        & \textbf{96.3}            \\
                                                                             & RESC       & 92.6        & 95.7        & \textbf{94.9}            \\ \bottomrule
\end{tabular}
\caption{Ablation studies of distance calculation method across different datasets. The best performance results are in \textbf{bold}.}
\label{tab:ab_dis}
\end{table*}

\begin{table*}[]
\centering
\begin{tabular}{@{}cccccc@{}}
\toprule
Task                                                                         & Dataset    & MeanShift     & DBSCAN        & Spectral Clustering & \textbf{\ Kmeans\ } \\ \midrule
\multirow{6}{*}{\begin{tabular}[c]{@{}c@{}}Image-level\\ (AUC)\end{tabular}} & MVTec-AD   & 96.3          & 97.0          & \textbf{97.8}                & \textbf{97.8}   \\
                                                                             & VisA       & 91.5          & 91.1          & \textbf{93.5}                & \textbf{93.5}   \\
                                                                             & MVTec LOCO & 68.7          & 70.4          & \textbf{71.8}       & 71.0            \\
                                                                             & BrainMRI   & 76.9          & 59.2          & 77.8                & \textbf{80.2}   \\
                                                                             & LiverCT    & 67.7          & 67.7          & 67.9                & \textbf{70.0}   \\
                                                                             & RESC       & \textbf{85.5} & \textbf{85.5} & \textbf{85.5}       & \textbf{85.5}   \\ \midrule
\multirow{6}{*}{\begin{tabular}[c]{@{}c@{}}Pixel-level\\ (AUC)\end{tabular}} & MVTec-AD   & 95.4          & 95.1          & 96.4                & \textbf{96.5}   \\
                                                                             & VisA       & 96.9          & 97.6          & \textbf{98.2}                & \textbf{98.2}   \\
                                                                             & MVTec LOCO & 73.1          & 74.0            & 74.5                & \textbf{75.1}   \\
                                                                             & BrainMRI   & 96.6          & 89.2          & 96.6                & \textbf{96.8}   \\
                                                                             & LiverCT    & 96.5 & 96.5          & \textbf{96.8}       & 96.3            \\
                                                                             & RESC       & \textbf{94.9} & \textbf{94.9} & \textbf{94.9}       & \textbf{94.9}   \\ \bottomrule
\end{tabular}
\caption{Ablation studies of clustering method across different datasets. The best performance results are in \textbf{bold}.}
\label{tab:ab_cluster}
\end{table*}

\section{Experimental Results in More Scenarios}
\label{app:E}

To further illustrate the versatility and robustness of UniVAD, we conduct experiments in various scenarios beyond standard anomaly detection datasets. These include real-world wood defect detection and crack segmentation tasks, both of which pose unique challenges and practical significance in industrial and structural inspection applications.

\subsection{Real-World Wood Defect Detection}

Wood, as one of the most commonly used and indispensable materials in industrial production, necessitates effective defect detection to ensure quality and reduce waste. To evaluate UniVAD's applicability in this domain, we collected a dataset comprising real-world wood samples from production environments and applied UniVAD for defect detection. The results, visualized in Figure~\ref{fig:more_visual} (a), demonstrate UniVAD's strong adaptability and generalization ability in this challenging real-world setting. 

\subsection{Crack Segmentation}

\begin{figure*}[t]
  \centering
   \includegraphics[width=0.95\textwidth]{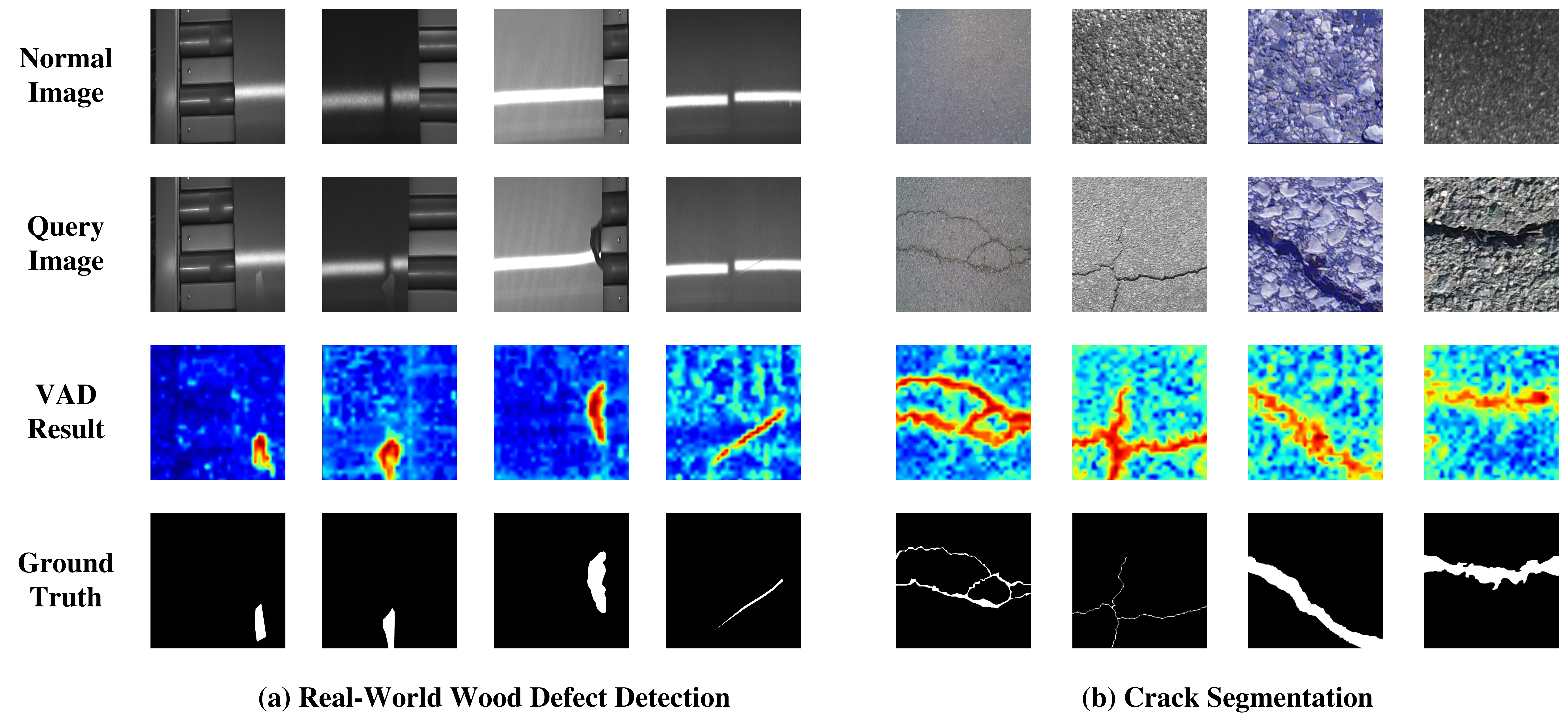}
   \caption{Visualization results in real-world wood defect detection and crack segmentation scenarios.}
   \label{fig:more_visual}
\end{figure*}

Crack segmentation, which involves detecting and delineating cracks on surfaces such as concrete, bricks, or other structural materials, is a critical task in applications like infrastructure maintenance and surface inspection. Given the significant implications for safety and cost-efficiency, effective methods for this task are highly valued. We evaluated UniVAD on a dedicated crack segmentation dataset, assessing its performance in few-shot scenarios where limited training data are available. As shown in Figure~\ref{fig:more_visual} (b), UniVAD achieves excellent segmentation accuracy on the CrackVision dataset, effectively identifying cracks even on complex and textured surfaces. These results further emphasize UniVAD's capability to address diverse and intricate anomaly detection tasks.

\end{document}